\documentclass[runningheads]{llncs}
\usepackage{caption}
\usepackage{multirow}
\usepackage{subcaption}
\usepackage{graphicx}
\usepackage{booktabs}
\usepackage{pgfplots}
\usepackage{floatrow}
\newfloatcommand{capbtabbox}{table}[][\FBwidth]
\usepackage{pgfplotstable}
\usepackage{filecontents}

\usepackage{longtable}

\usepackage{amsmath}
\usepackage{mathtools}
\usepackage{cite}
\usepackage{algorithm}
\usepackage{{algpseudocode}}

\definecolor{Blue1}{RGB}{0, 2, 92}
\definecolor{Blue2}{RGB}{151, 0, 94}
\definecolor{Blue3}{RGB}{255, 166, 0}
\definecolor{Blue4}{RGB}{232, 60, 63}
\definecolor{Blue5}{RGB}{90, 90, 90}

\definecolor{Green1}{RGB}{0, 92, 78}
\definecolor{Green2}{RGB}{57, 130, 67}
\definecolor{Green3}{RGB}{144, 161, 28}
\definecolor{Green4}{RGB}{255, 174, 0}

\definecolor{Purp1}{RGB}{24, 125, 153}
\definecolor{Purp2}{RGB}{0, 158, 136}
\definecolor{Purp3}{RGB}{118, 177, 60}
\definecolor{Purp4}{RGB}{255, 166, 0}

\usepackage{pstricks}

\usepackage{enumitem,amssymb}
\newlist{todolist}{itemize}{2}
\setlist[todolist]{label=$\square$}
\usepackage{pifont}
\newcommand{\cmark}{\ding{51}}%
\newcommand{\xmark}{\ding{55}}%

\newcommand{\etal}{et al.~}
\newcommand{\ie}{\textit{i}.\textit{e}.~}
\newcommand{\eg}{e.g.~}

\usepackage{graphicx}

\begin{document}
\title{L2PF - Learning to Prune Faster
%\thanks{Supported by BMW AG.}
}

\author{Manoj-Rohit Vemparala *\inst{1},
Nael Fasfous* \inst{2}, Alexander Frickenstein*\inst{1},\\
Mhd Ali Moraly* \inst{1}, Aquib Jamal \inst{1}, Lukas Frickenstein \inst{1}, \\
Christian Unger \inst{1} Naveen-Shankar Nagaraja \inst{1} Walter Stechele\inst{2}\\
* indicates equal contributions}
% \author{\IEEEauthorblockN{Alexander Frickenstein\textsuperscript{1\**}, Manoj-Rohit Vemparala\textsuperscript{1\**}, Nael Fasfous\textsuperscript{2\**}, Laura Hauenschild\textsuperscript{2\**},\\
% Naveen-Shankar Nagaraja\textsuperscript{1}, Christian Unger\textsuperscript{1}, Walter Stechele\textsuperscript{2}}
% \IEEEauthorblockA{\textsuperscript{1}\textit{Autonomous Driving, BMW Group,  Munich, Germany} \\
% \textsuperscript{2}\textit{Department of Electrical and Computer Engineering, Technical University of Munich, Munich, Germany}\\
% \textsuperscript{1}\{$<\mathtt{firstname}>$.$<\mathtt{lastname}>$\}@bmw.de, \textsuperscript{2}\{$<\mathtt{firstname}>$.$<\mathtt{lastname}>$\}@tum.de
\institute{BMW Autonomous Driving \and Technical University of Munich\\
\textsuperscript{1}$\mathtt{firstname.lastname@bmw.de}$, \textsuperscript{2}$\mathtt{firstname.lastname@tum.de}$}

\maketitle             
% \input{misc/todoes.tex}

%\setlength{\belowcaptionskip}{-10ex}
%            ____   _____ _______ _____            _____ _______ 
%      /\   |  _ \ / ____|__   __|  __ \     /\   / ____|__   __|
%     /  \  | |_) | (___    | |  | |__) |   /  \ | |       | |   
%    / /\ \ |  _ < \___ \   | |  |  _  /   / /\ \| |       | |   
%   / ____ \| |_) |____) |  | |  | | \ \  / ____ \ |____   | |   
%  /_/    \_\____/|_____/   |_|  |_|  \_\/_/    \_\_____|  |_|   
\begin{abstract}
Various applications in the field of autonomous driving are based on convolutional neural networks (CNNs), especially for processing camera data. The optimization of such CNNs is a major challenge in continuous development. Newly learned features must be brought into vehicles as quickly as possible, and as such, it is not feasible to spend redundant GPU hours during compression. In this context, we present Learning to Prune Faster which details a multi-task, try-and-learn method, discretely learning redundant filters of the CNN and a continuous action of how long the layers have to be fine-tuned. This allows us to significantly speed up the convergence process of learning how to find an embedded-friendly filter-wise pruned CNN. For ResNet20, we have achieved a compression ratio of $3.84 \times$ with minimal accuracy degradation. Compared to the state-of-the-art pruning method, we reduced the GPU hours by $1.71 \times$.  
\end{abstract}
\label{sec:abstract}

%  _____ _   _ _______ _____   ____  _____  _    _  _____ _______ _____ ____  _   _ 
% |_   _| \ | |__   __|  __ \ / __ \|  __ \| |  | |/ ____|__   __|_   _/ __ \| \ | |
%   | | |  \| |  | |  | |__) | |  | | |  | | |  | | |       | |    | || |  | |  \| |
%   | | | . ` |  | |  |  _  /| |  | | |  | | |  | | |       | |    | || |  | | . ` |
%  _| |_| |\  |  | |  | | \ \| |__| | |__| | |__| | |____   | |   _| || |__| | |\  |
% |_____|_| \_|  |_|  |_|  \_\\____/|_____/ \____/ \_____|  |_|  |_____\____/|_| \_|
\section{Introduction}
\label{sec:introduction}
    With the advent of scalable training hardware and frameworks, the trend towards training larger deep neural networks (DNNs) or ensembles of networks has become more prevalent than ever~\cite{he2015deep}. As a result, compression of DNNs has become an increasingly popular field of research in recent years. This is particularly the case for convolutional neural networks (CNNs), which have become the state-of-the-art solution for most computer vision problems, and often find applications in embedded scenarios, necessitating a reduction in their storage requirements and computational costs for inference. In the field of autonomous driving~\cite{bdad}, embedded hardware~\cite{orthrus,WinoCNN} is highly constrained and short development cycles are key for being first to market.

Many standard techniques exist in literature to reduce the number of network parameters and the complexity of the computations involved in CNNs~\cite{he2015deep, He17ChannelPr}. Due to their inherent complexity and redundancy, CNNs can sustain many forms of structural and algorithmic approximation while still delivering adequate functional accuracy w.r.t. the given task, \eg image classification. With the help of fine-tuning iterations, CNNs can recover the lost accuracy after a compression method has been applied, with negligible degradation.

Network compression can be viewed as a standard optimization problem. The search space composes of the possible combinations to compress neurons, kernels or layers. 
%The number of possible combinations to compress neurons, kernels or layers in a CNN and to which extent, induce a large and complicated network design space.
% \lukas{The number of possible combinations to compress neurons, kernels or layers in a CNN and to which extent, induce a large and complicated network design space}. 
Exploring this design space is a difficult task when considering its size and the computational overhead required to sufficiently evaluate each potential configuration in it. %In the context of CNNs, the removal of any neuron, presents a unique solution for element-wise pruning. More coarsely, pruning at the kernel level, reduces the size of the search space significantly, at the cost of possibly obscuring the global optima from being reached. Scaling up to filter-wise and/or channel-wise pruning amplifies the effect of this trade-off.
Referring back to the healing effect of network fine-tuning after pruning, efficiently traversing this search space necessitates a fair balance between the number of epochs the fine-tuning is done for and the computational overhead required by those fine-tuning epochs. This ultimately leads to shorter product and time-to-market cycles and facilitates continuous development.

As with many other optimization problems, existing literature covers various forms of design space exploration w.r.t. pruning~\cite{He2018AMCAF,Huang2018LearningTP,Netadapt}. Automated pruning was an inevitable step in the evolution of this compression technique due to its complex nature. Automated pruning makes tools usable for researchers and product developers with little to no background in CNN optimization.
% (expert knowledge). %\alex{automation->makes pruning usable as a tool which is usable by researcher and product developers with no optimization background (expert knowledge)}
%Many works have leveraged reinforcement learning (RL) agents to help in solving this problem~\cite{Netadapt, He2018AMCAF}, while others have integrated pruning directly into the training phase~\cite{Frickenstein2020ALFAL, StructADMM}. Among automated pruning techniques, the decision criteria, based on which a neuron is pruned or kept, can vary. An automated approach combined with a heuristic-based decision criteria may end up constraining solutions to those which conform only with the effectiveness of the heuristic~\cite{He2018AMCAF}. Other decision criteria, which may resolve to an equivalent solution in terms of network accuracy, can additionally benefit in other domains by understanding the behaviour of the network. This allows an automated pruning technique to consciously remove elements based on a richer set of guiding information, potentially considering other critical criteria such as network robustness and feature-awareness.

In this work, we build upon a learning-based pruning approach~\cite{Huang2018LearningTP} to tackle the challenge of choosing the optimal number of fine-tuning epochs~\footnote{An epoch describes a complete cycle of a data set, \ie training data set.} for a potential pruning solution. The decisions of our RL-based pruning agent are based not only on the features embedded in the CNNs kernels, but also on the fine-tuning potential of the individual layers. This results in feature-conscious decisions % which can potentially improve the robustness of the resulting network against malicious adversarial attacks 
and reduced overall time required by the pruning technique.
%\lukas{Reference? Or do we prove it?}\nael{I think we can hint at it with the CAM section, but we dont have time to prove it unfortunately}.
The contributions of this paper can be summarized as follows:

\begin{itemize}
    \item A multi-task learning approach involving a reinforcement learning agent, which learns a layer's features and adequate fine-tuning time concurrently.
    \item Formalizing the design space exploration problem w.r.t. pruning effectiveness and time-effort.
    \item A study on the sequence of layer-wise pruning of a convolutional neural network.
    % \item Robustness evaluation on different automated pruning techniques
\end{itemize}

%   _____  ______ _            _______ ______ _____   __          ______  _____  _  __
%  |  __ \|  ____| |        /\|__   __|  ____|  __ \  \ \        / / __ \|  __ \| |/ /
%  | |__) | |__  | |       /  \  | |  | |__  | |  | |  \ \  /\  / / |  | | |__) | ' / 
%  |  _  /|  __| | |      / /\ \ | |  |  __| | |  | |   \ \/  \/ /| |  | |  _  /|  <  
%  | | \ \| |____| |____ / ____ \| |  | |____| |__| |    \  /\  / | |__| | | \ \| . \ 
%  |_|  \_\______|______/_/    \_\_|  |______|_____/      \/  \/   \____/|_|  \_\_|\_\
\section{Related Work}
\label{sec:related_work}
%Neural network compression techniques such as quantization~\cite{haq,dorefa}, decomposition~\cite{Frickenstein2020ALFAL} and factorization~\cite{JaderbergVZ14} or pruning \cite{He2018AMCAF,Huang2018LearningTP,StructADMM} are identified as key methods to mitigate discrepancies between the constrained nature of embedded systems and the ever increasing resource demands of deep neural networks - all pursuing to develop more efficient CNNs for deployment on embedded hardware.
In the following sections, we classify works which use pruning techniques into heuristic-based, in-train and learning-based strategies.%
%Table
\iffalse
    \begin{table}[t]
    \caption{Classification of CNN pruning methods.}
    %\vspace{-8mm}
    \centering
    \resizebox{0.8\textwidth}{!}{
    \begin{tabular}{l|cc|cccc}
    \toprule
    % &\multicolumn{2}{c|}{Heuristics}&\multicolumn{3}{ccc}{Automated}\\
    
    &\textbf{Irregular}& \textbf{Regular} & \textbf{In-train} &\textbf{AMC} & \textbf{Try\&Learn}   & \textbf{L2PF} \\
    \textbf{Advantage}&\cite{geometric_mean_filter} 
    &\cite{Huang2018LearningTP, He17ChannelPr} &\cite{Frickenstein2020ALFAL, StructADMM}& \cite{He2018AMCAF}
    %\cite{ Netadapt,Tan2018MnasNetPN,cai2018proxylessnas,He2018AMCAF}
    %&\cite{timeloop,EnergyAwarePruning}&[Ours]\\
    & \cite{Huang2018LearningTP} & [Ours]\\
    \midrule
    \midrule
    HW benefits: &\xmark& \cmark & \cmark& \cmark & \cmark& \cmark\\ 
    Learning based: &\xmark&\xmark& \cmark&\cmark&\cmark& \cmark\\
    Granular search:  &\xmark&\xmark& \cmark&\xmark&\cmark& \cmark\\
    Exploration GPU hours: & \xmark& \xmark& \cmark& \cmark&  \xmark& \cmark\\
    %Expertise independent: & \xmark& \xmark& \xmark& \xmark&  \xmark& \cmark\\
    \bottomrule
    \end{tabular}}
    \label{tab:related_work}
    \end{table}
    %\vspace{-7ex}
\fi
\subsubsection{Heuristic-based Pruning:}
% Characteristic
    Heuristic-based compression techniques consider static or pseudo-static rules that define the compression strategy, when pruning an underlying CNN. 
    % The saliency of neurons is determined by the hand-crafted heuristics. 
% Related Work
% Irregular Pruning
    The pruning method proposed by~\cite{learn_weights} utilizes the magnitude of weights, where values below a threshold identify expendable connections. Such pruning of individual weights, as presented in~\cite{learn_weights}, leads to inefficient memory accesses, rendering irregular pruning techniques impractical for most general purpose computing platforms. 
% Regular Pruning
    Regularity in pruning is a key factor in accelerator-aware optimization. Frickenstein et al.~\cite{Frickenstein_2019_CVPR_Workshops} identifies redundant kernels in the weight matrix based on magnitude based heuristic. He et al.~\cite{geometric_mean_filter} prune redundant filters based on the geometric median heuristic of the filters. In~\cite{He17ChannelPr}, He et al. introduce an iterative two-step algorithm to effectively prune layers in a given CNN. First, redundant feature maps are selected by LASSO regression followed by minimizing the output errors of the remaining feature maps by solving least squared minimization. 
    %\newline  \lukas{Add RAO Paper} \newline  
%Drawbacks
    %Despite the mentioned hardware benefits of regular pruning, both regular and irregular rule-based pruning require domain expertise to formalize suitable heuristics and pruning rules, where the variety of CNNs in complexity, structure and target task lead to the challenge of finding one-size-fits-all rules involving different compression criteria. Additionally, the severe reduction of the search space may result in a poor choice of design parameters, resulting in sub-optimal solutions.  
    %\begin{itemize}
    %    \item Con: Required domain expertise
    %    \item Con: Sub-optimal solution
    %    \item Con: Model dependent
    %\end{itemize}
    \vspace{-3ex}
\subsubsection{In-train Pruning:}
% Characteristic
    Integrating the pruning process into the training phase of CNNs is characterized as in-train pruning. 
    % Such methods utilize the gradients of a given cost function beyond the standard procedure to update the parameters of a respective network, but also to update the pruning masks based on these gradients.
% Related Work
% ALF
    The auto-encoder-based low-rank filter-sharing technique (ALF) proposed by Frickenstein et al.~\cite{Frickenstein2020ALFAL} utilizes sparse auto-encoders that extract the most salient features of convolutional layers, pruning redundant filters. %ALF approximates weight filters of existing CNNs and thus, reduces the gap between increasing hardware requirements of state-of-the-art networks and the constrained setup of embedded applications.
% ADMM
    %Zhang et al.~\cite{} present a systematic weight pruning framework for DNNs, where the weight pruning problem is formulated as a constrained  non-convex optimization problem subject to constraints on the cardinality of weights in each layer, resulting in a high degree of sparsity by incorporating hard constraints on the weights. By leveraging alternating direction method of multipliers (ADMM), the optimization problem can be decomposed into two subprolbems which then are solved separately 
% StructADMM
    Zhang et al.~\cite{StructADMM} propose ADAM-ADMM, a unified, systematic framework of structured weight pruning of DNNs, that can be employed to induce different types of structured sparsity based on ADMM. 
% % Pro/Con
%     \nael{Could be removed: When using structured pruning in the training scope, the feasibility of pruned and more efficient CNNs is improved. Furthermore, the presented in-train methods provide a learning-based pruning scheme capable of learning salient features during the optimization. 
%     % Question: What expertise is required for Alf? 
%     However, the lack of GPU-hour awareness is a limitation of recent pruning methods that incorporate the pruning process into the training phase of CNNs.}\lukas{Use this space !}
\vspace{-3ex}
\subsubsection{Learning-based Pruning:}
% Characteristic
Defining the pruning process as an optimization problem and exposing it to an RL-agent has been done in a variety of works~\cite{He2018AMCAF, Huang2018LearningTP}.
% Related Work
% AMC
He et al.~\cite{He2018AMCAF} demonstrate a channel pruning framework leveraging a Deep Deterministic Policy Gradient (DDPG) agent. Their framework first learns the sparsity ratio to prune a layer, while actual channel selection is performed using $\ell1$ criteria. 
%Pro/Con:
The appealing higher compression ratio, better preservation of accuracy as well as faster, coarse and learnable exploration of the design space with few GPU hours, highlight the strong points of the AMC framework. However, this leaves room for improvement in search-awareness, as AMC is unaware of the exact features it is pruning.
%\lukas{@Manoj/Nael: could you verify: since its stated contrary in the table}\nael{fixed in table}
% L2P
Huang et al.~\cite{Huang2018LearningTP} demonstrated a 'try-and-learn' RL-based filter-pruning method to learn both sparsity ratio and the exact position of redundant filters, but it leaves out the number of fine-tuning epochs as a hyper-parameter. Here, the optimal value can change, depending on the model's architecture and the data set at hand.
We extend the work of Huang et al.~\cite{Huang2018LearningTP} incorporating GPU hour awareness by proposing an iterative method with a learned minimum number of iterations for fine-tuning, hyper-parameter free method. 
% This try-and-learn method automatically learns the pruned architecture by finding the exact position of filters without requiring any pre-determined sparsity ratio, sensitivity analysis or heuristic pruning criteria.
% To the best of our knowledge, this RL-based 'try-and-learn' pruning method is the first aiming to balance highest possible compression ratio, while maintaining accuracy \textit{and} identifying the minimum number of retraining epochs of the respective pruned layers to consider with respect to its retraining potential. 

%   _____  _        _____  _____  _    _ _   _ ______ 
%  |  __ \| |      |  __ \|  __ \| |  | | \ | |  ____|
%  | |__) | |      | |__) | |__) | |  | |  \| | |__   
%  |  _  /| |      |  ___/|  _  /| |  | | . ` |  __|  
%  | | \ \| |____  | |    | | \ \| |__| | |\  | |____ 
%  |_|  \_\______| |_|    |_|  \_\\____/|_| \_|______|
\section{Method}
\label{sec:method}
In this chapter, we propose a multi-task learning approach, namely \emph{Leaning to Prune Faster} (L2PF) involving a RL-agent, which learns a layer's features and adequate fine-tuning time concurrently.
In this regard, the pruning problem in the context of a RL framework is formulated in Sec.~\ref{sec:formulation}. The environment and state space is defined in Sec.~\ref{sec:env}. We discuss the discrete and continuous action spaces in Sec.~\ref{sec:action_space}, and the reward formulation in Sec.~\ref{sec:reward}. Lastly, the agent's objective function is formulated in Sec.~\ref{sec:objective_function}.

    \subsection{Problem Formulation}
    \label{sec:formulation}
\begin{figure}[h]
    \centering
    \includegraphics[width=0.8\textwidth]{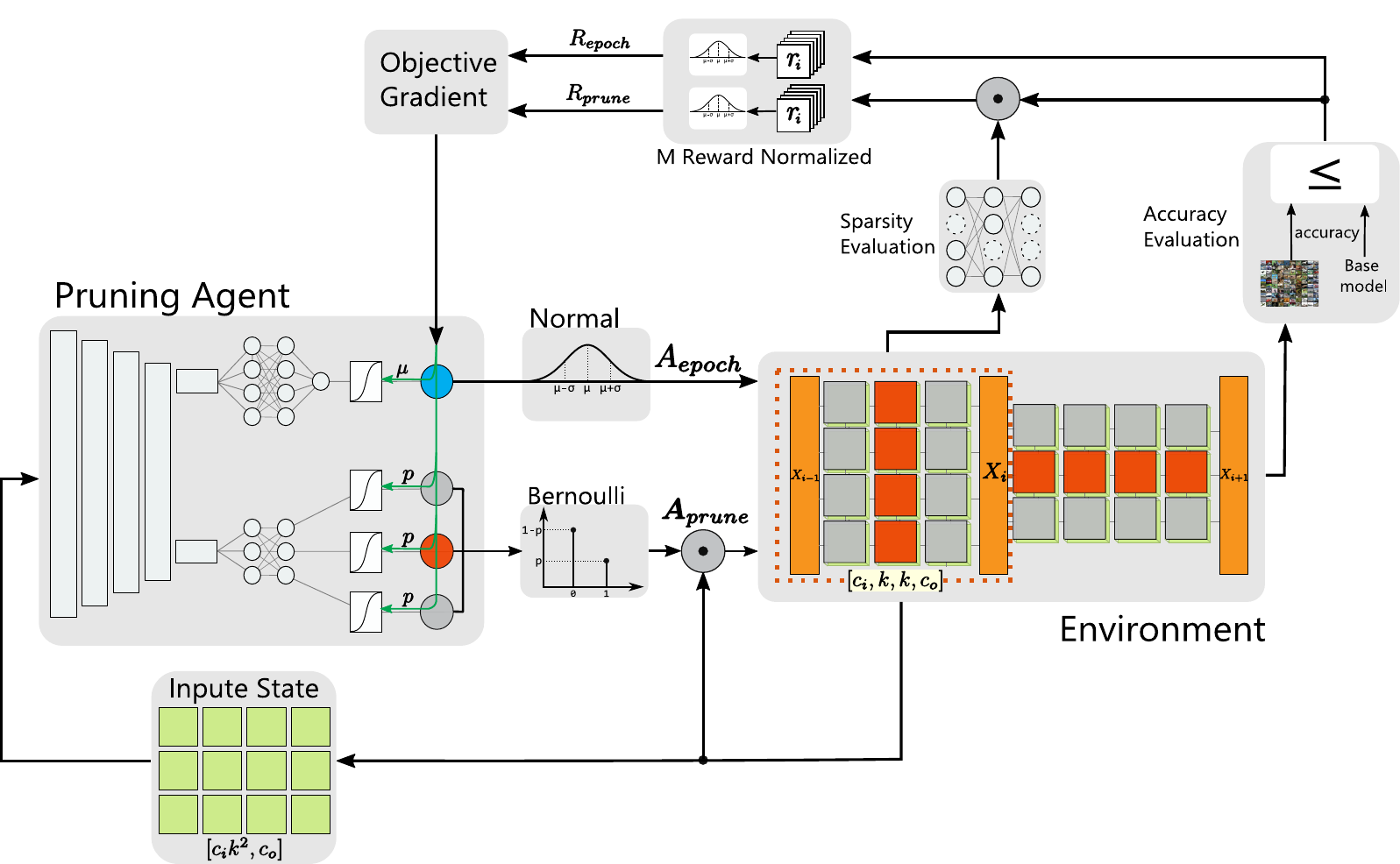}
    \caption[Agent-environment dynamics]{Agent receives rewards and weights as input state, whereas environment receives both prune and epoch actions. In each prune episode, M=5 Monte-Carlo set of actions are sampled ($\mathbf{A}_\text{prune}$ and $A_\text{retrain}$). The corresponding $M$ rewards ($R_\text{prune}$ and $R_\text{retrain}$) are normalized to zero mean and unit variance~\cite{Huang2018LearningTP}.} 
    % \manoj{Remove border}
    \label{fig:env_agent}
\end{figure}

    The structured filter-pruning task within an RL framework can be expressed as a `try-and-learn' problem, similar to the work from Huang et al.~\cite{Huang2018LearningTP}. We aim to find the best combination of filters that achieve the highest compression ratio (CR) while incurring a minimum loss of accuracy (Acc) and requiring a minimum number of fine-tuning epochs during the exploration episodes. Fig.~\ref{fig:env_agent} demonstrates the interplay between the proposed pruning agent and CNN environment. The proposed method is able to learn three aspects: First, the minimum number of epochs required to explore each pruning strategy. Second, the degree of sparsity of each layer in the model. Third, the exact position of the least important filters to be pruned. 

Formally, let $B$ be a fully-trained model with $L$ layers and the input of the $\ell^{th}$ convolutional layer has a shape $[ {c}^\ell \times {w}^\ell \times {h}^\ell]$, 
% \lukas{Rework that it fits in one line}
% $\left[ {c}^\ell \times {w}^\ell \times {h}^\ell \right]$ 
where $c^\ell$, ${h}^\ell$ and ${w}^\ell$ represents number of input channels, height and width. 
The $\ell^{th}$ layer is convolved with the weight tensor $\mathbf{W}^\ell$, \ie 2D convolutional layer's trainable parameters, with shape $\left[ N^\ell \times c^\ell \times {k}^\ell \times {k}^\ell \right]$, where ${k}^\ell$ represents the kernel size and  $N^\ell$ is number of filters. 
After pruning $n^\ell$ filters, the weight tensor is of shape $\left[ ({N^\ell-n^\ell}) \times c^\ell \times {k}^\ell \times {k}^\ell \right]$. 
To enable a direct comparison with the work of He et al.~\cite{He2018AMCAF}, the layer compression ratio is defined as $\frac{c^\ell-n^{\ell-1}}{c^\ell}$. 
Additionally, we define model compression ratio to be the total number of weights divided by the number of non-zero weights. 
% \manoj{Is this compression ratio of the entire model?}
% \lukas{We need to clarify: Layer sparsity ratio, model compression ratio, ...}
% \ali{re-formulate more clearly}
% model sparsity ratio, computed as the the number of zero-valued weights divided by the number of all weights as well as the model compression ratio, calculated by the total number of weights divided by the number of non-zero weights, are adopted accordingly.

%To enable direct comparison with the work of He et al. in~\cite{He2018AMCAF}, we also define layer sparsity ratio as $\frac{m^\ell-n^{\ell-1}}{m^\ell}$ (lower is better), while model sparsity ratio is defined to be $\frac{\#zero\_weights}{\#total\_weights}$, and model compression ratio is $\frac{\#total\_weights}{\#nonzero\_weights}$ (higher is better). 
%\lukas{Alternative - Do not use formulas, rather describe it in a sentence like in AMC paper. To enable a direct comparison to the work of He et al.~\cite{He2018AMCAF}, the model sparsity ratio, defined as the the number of zero-valued weights divided by the number of all weights, as well as the model compression ratio, calculated by the total number of weights divided by the number of non-zero weights, are adopted accordingly.}

    \vspace{-7ex}
    \subsection{Environment}
    \label{sec:env}
    The environment is the pre-trained CNN model $B$ to be pruned. The \texttt{state space} is the fully trained weight tensor $\mathbf{W}^\ell$ of the layer to be pruned, which is used as an input for the agent, similar to Huang et al.~\cite{Huang2018LearningTP}. 
% \ali{mention the sahpe of agent input} 
%which has one state namely,
For each layer (or residual block), a new agent is trained from scratch. The environment receives two actions from the agent: pruning action $\mathbf{A}_\text{prune}$ and fine-tuning epoch action ${A}_\text{retrain}$. Subsequently, it generates a reward $R = R_\text{prune} + R_\text{retrain}$. For each filter there is a binary mask ${\mathbf{m}}_{i}^{\ell}\in{\left\{0,1\right\}}^{{c}^\ell \times {k}^\ell \times {k}^\ell}$. Pruning the $i^{th}$ filter $\mathbf{W}_{i}^\ell$ in layer $\ell$ is performed by element-wise multiplication between the filter $\mathbf{W}_{i}^\ell$ and its corresponding mask $\mathbf{m}_i^\ell$ . When pruning $\mathbf{W}_{i}^\ell$, the $i^{th}$ kernel of all filters in the ${(\ell+1)}^{th}$ layer are also pruned. At each pruning step, masks are updated according to $\mathbf{A}_\text{prune}$ and the environment is fine-tuned for a few epochs %% \nael{can be made into a greek symb? or simply ${e}_\text{retrain}$}
${e}_\text{retrain}$.
% the corresponding $i^{th}$ kernels from the ${(\ell+1)}^{th}$ layer's filter are pruned, consequently 
    \subsection{Distinct Action Space for Pruning and Epoch-Learning}
    \label{sec:action_space}
    \label{sec:action_space}
The action space of the proposed RL-framework is split into two distinct spaces to satisfy the discrete and continuous requirements of actions for pruning and epoch learning respectively.
\vspace{-3ex}
\subsubsection{Discrete Pruning Action Space:}
The discrete pruning action space is the combination of all possible prune actions $\mathbf{A}_\text{prune}$. It is clear that action space dimension grows exponentially as $\mathcal{O}(2^N)$, where $N$ is the number of filters in a layer.
Discrete actions are sampled from $N$ independent stochastic Bernoulli units~\cite{williams1992}. Each unit has one learnable parameter $p$ that represents the probability of keeping the filter. 
%% \manoj{Verify with learning to prune paper} 
\vspace{-3ex}
\subsubsection{Continuous Epoch-Learning Action Space:}
The continuous epoch-learning is used to determine the number of fine-tuning epochs ${e}_\text{retrain}$. 
Like the discrete action, the continuous action ${A}_\text{retrain}$ must also be sampled from some distribution. 
Practically, ${e}_\text{retrain}$ takes values within a bounded range $\in \mathbb{Q}^{+}$. 
A continuous action ${A}_\text{retrain}$ is sampled from a Normal distribution which has two learn-able parameters $\mu$ and $\sigma$. Since ${e}_\text{retrain}$ is bounded while Normal distribution has unbounded support, it must be truncated. 
Truncating a Normal distribution might cause the estimated policy to get biased into the direction of the truncation boundary where the reward peaks (boundary effect)~\cite{chou17a}. 
To circumvent the boundary effect, we employ the approach from Chou et al.~\cite{chou17a}. 
The sampled action ${A}_\text{retrain}$ is sent to the environment with no alteration. 
To calculate ${e}_\text{retrain}$, %%~\nael{the environment doesn't truncate, its the CNN. We can say the action is truncted (passive)}
the action value is truncated within $[0, 1]$. However, for gradient calculations, non-truncated action values are used. 

    \subsection{Multi-objective Reward Function}
    \label{sec:reward}
    The quality of agent action is conveyed back to the agent by the reward signal, $R_\text{prune}$ and $R_\text{retrain}$ for $\boldsymbol{A}_{prune}$ and $A_{retrain}$ respectively.

\subsubsection{Prune Reward:} 
The prune reward $R_\text{prune}$ is a measure for sparsity level and model accuracy ${acc}_\text{pruned}$. It promotes actions that remove filters with minimum accuracy loss of the pruned model w.r.t. the validation set. Following the work of Huang et al.~\cite{Huang2018LearningTP}, we define the prune reward as a product of two terms, \ie  ${acc}_\text{term}$ and ${eff}_\text{term}$, as stated in Eq.~\ref{eq:prune_reward}. 
\begin{equation}
    \label{eq:prune_reward}
    {R}_{\text{prune}}\left(\mathbf{A}^\ell_{\text{prune}}, {acc}_\text{pruned} \right) = {acc}_\text{term} \cdot {eff}_\text{term}
\end{equation}
    
\subsubsection{Accuracy Term: }
% ${acc}_\text{term}$ is used to evaluate the effect of $\mathbf{A}_\text{prune}$ with respect to the accuracy. It is measured after retraining environment for few epochs according to $A_\text{retrain}$. 
Similar to Huang et al.~\cite{Huang2018LearningTP}, ${acc}_\text{term}$ is defined in Eq.~\ref{eq:acc_term}.
% \lukas{Prune: (Eq. (1) already introduced) Similar to Huang et al.~\cite{Huang2018LearningTP}, ${acc}_\text{term}$ is defined in Eq.~\eqref{eq:acc_term}.} 
% \ali{Done} 
The bound $b$ is a hyper-parameter introduced in the reward function to allow control over the trade-off between model compression and tolerable accuracy drop.
When the accuracy drop is greater than $b$, ${acc}_\text{term}$ is negative, otherwise it lies in the range $[0, 1]$ respectively 0 and 100\%.
    
% \lukas{Can be boiled down - When no accuracy loss is incurred then ${acc}_\text{term}$ is one, when accuracy is lower than $b$ then it is negative. Otherwise it is in range ${acc}_\text{term} \in [0, 1]$.} \ali{Done}
\begin{equation}
    \label{eq:acc_term}
    {acc}_\text{term} = \frac{b-\max\left[{0, {acc}_\text{base} - {acc}_\text{pruned}}\right]}{b}
\end{equation}

%% In case of searching for the most efficient pruned model with no loss of accuracy, Eq\eqref{eq:acc_term_no_bound} is utilized.  

%%\begin{equation}
%%    \label{eq:acc_term_no_bound}
%%    {acc}_{term} = \begin{dcases}
%%    1 & \text{if ${acc}_{fully\_trained}={acc}_{pruned}$} \\
%%    {acc}_{pruned}-{acc}_{full\_trained} & \text{otherwise}
%%    \end{dcases}
%%\end{equation}

\subsubsection{Efficiency Term: } 
% ${eff}_\text{term}$ measures the induced sparsity level.  Similar to~\cite{Huang2018LearningTP}, ${eff}_\text{term}$ is defined in Eq.~\eqref{eq:eff_term}. 
To prevent the agent from changing the model depth, the efficiency term ${eff}_\text{term}$ proposed by Huang et al.~\cite{Huang2018LearningTP} is extended as shown in Eq.~\ref{eq:eff_term}. 
If the prune action is aggressive, the accuracy drop will be less than the bound $b$ resulting in a negative reward. 
If layer sparsity ratio is low, ${eff}_\text{term}$ will drive reward to zero.
% \lukas{Check all equation references - comment out only after all equation references are equal.} 
% \lukas{Same pattern as for acc term, can be boiled down or delet: If no filter is pruned then it is zero. If all filters are pruned ${eff}_\text{term}$ is -1 preventing agent from changing model depth. For all other cases ${eff}_\text{term}$ has positive value between $[0, \log (\text{\#total\_filters})]$} \ali{Done}
%\begin{equation}
%    \label{eq:eff_term}
%    {eff}_\text{term} = \begin{dcases}
%        \log{\frac{\text{\#total\_filters}}{\text{\#remain\_filters}}} & \text{if \hspace{0.25cm} \#remain\_filters $\leq$ \#total\_filters} \\
%                    -1 & \text{if \hspace{0.25cm} \#remain\_filters $=$ 0}
%        \end{dcases}
%\end{equation}
\begin{equation}
    \label{eq:eff_term}
    {eff}_\text{term} = \begin{dcases}
        \log{\frac{N}{(N-n)}} & \text{if} \hspace{0.25cm} (N-n) \leq N \\
                    -1 & \text{if} \hspace{0.25cm} (N-n) = 0
        \end{dcases}
\end{equation}

\subsubsection{Fine-tuning Epoch Reward:}
The fine-tuning epoch reward $R_\text{retrain}$ is responsible for promoting a lower number of fine-tuning epochs. The reward is expressed in Eq.~\ref{eq:retrain_reward}. An action is considered \emph{good} when $ \lvert A_\text{retrain} \rvert$ is low without causing an intolerable accuracy drop. If the environment incurs no accuracy loss then $R_\text{retrain}=0$, when loss is incurred then it will be a negative value scaled by the absolute value of $A_\text{retrain}$.

\begin{equation}
    \label{eq:retrain_reward}
    R_\text{retrain}\left({A}_\text{retrain}, {acc}_\text{pruned} \right) = \lvert A_\text{retrain} \rvert \times \left({acc}_\text{pruned}-{acc}_\text{base}\right)
\end{equation}
In each prune episode, $M$ Monte-Carlo set of actions are sampled again resulting in $M$ corresponding rewards $R_\text{prune}$ and $R_\text{retrain}$. 
The reward values are normalized to zero mean and unit variance for both set of rewards~\cite{hasselt2016learning, mohamed2019monte}.

    \subsection{Agent Design}
    \label{sec:objective_function}
    The agent is a non-linear stochastic functional approximator parameterized by $\mathbf{\theta}$. It is composed of four convolutional layers, two classifiers each with two feed-forward layers~\cite{Huang2018LearningTP}, and two types of stochastic output units, \ie Bernoulli and Normal. 
The agent parameters are $\mathbf{\theta}={\{\mathbf{w}, \mu, \mathbf{P}\}}$,
where parameters $\mathbf{w}$ 
%%\lukas{@Ali: verify if \mathbf{w} or \mathbf{W}}
are  the agent weights, $\mu$ is a learnable parameter to sample the fine-tuning action ${A}_\text{retrain}$, and $\mathbf{P}$ is the set of probabilities for Bernoulli units. 
The agent outputs two actions: discrete action $\mathbf{A}_\text{prune}$ for pruning, and continuous action ${A}_\text{retrain}$ for fine-tuning epochs.
The \texttt{pruning action}  $\mathbf{A}_\text{prune}$ is a set $\{a_{1}^{\ell}, a_{2}^{\ell}, ... ,a_{N^l}^{\ell}\}$, where $a_{i}^{\ell} \in \{0, 1\}$ is equivalent to $\{prune, keep\}$ and $N^\ell$ is the number of filters in the $\ell^{th}$ layer~\cite{Huang2018LearningTP}. Using this scheme, the agent is able to explore both sparsity ratio and to select the exact position of filters to prune.

% \sussubsection{Fine-tuning Action:} 
The \texttt{fine-tuning action} ${A}_\text{retrain}$ is a continuous action sampled from a normal distribution with two parameters - $\mu$, $\sigma$. The mean $\mu$ is a learnable parameter, while $\sigma$ is chosen to be non-learnable and set to be proportional to $\lvert R_\text{retrain} \rvert$~\cite{williams1992, Sutton1999}. 
The value of $\sigma$ controls how far a sample can be from the mean. 
When reward signal $R_\text{retrain}$ is low indicating bad actions, then $\sigma$ takes higher value which allows the agent to explore actions further away from $\mu$. 
Actions ${A}_\text{retrain} \notin [0, 1]$ are 
considered \emph{bad} and give negative reward. The environment fine-tunes for the number of epochs given in Eq.~\ref{eq:epoch_retrain}, where $\beta$ is an upper limit for $e_\text{retrain}$.
    
    \begin{equation}
    \label{eq:epoch_retrain}
         e_\text{retrain} = \min[\max [0, {A}_\text{retrain}], 1] \times \beta
    \end{equation}
    
    % equals to ${truncated\_A}_\text{retrain} \times max\_retrain\_epochs$, where ${truncated\_A}_\text{retrain}$ is calculated by the environment using the formula $\min[\max [0, {A}_\text{retrain}], 1]$. 
    % \lukas{find more convenient way of expressing. } \ali{Done}
    
%% \lukas{which reward signal? The overall signal R or the respective $R_{epochs}$}
    % \item \texttt{Retraining action} ${epoch}_{retrain}$ is continuous action, sampled from a normal distribution with two parameters, the mean $\mu$ is a learnable parameter, while $\sigma$ is chosen to be non-learnable and set to be proportional to $1-R_{epochs}$. The value of $\sigma$ controls how far a normal sample is away from the average. When reward signal \lukas{which reward signal? The overall signal R or the respective $R_{epochs}$ } is low indicating bad actions, then $\sigma$ takes higher value which allows the agent to sample and explore actions further away from the average. The environment retrains for a number of epochs equals to ${truncated\_epoch}_{retrain} \times max\_retrain\_epochs$. The ${truncated\_epoch}_{retrain}$ is calculated by the environment using the formula 

%% to make shorter and add into Agent subsection
We leverage the stochastic policy gradient (SPG) method to find an optimal policy $\pi^*$. SPG is guaranteed to converge at least to a local optimum without requiring the state space distribution~\cite{sutton2018reinforcement}. Our objective function $J(\mathbf{\theta})$ is the expected sum of all rewards over one episode. The objective gradient w.r.t. the policy parameters is given in Eq.~\ref{eq:Obective_grad}. Both terms in Eq.~\ref{eq:Obective_grad} can be solved approximately using the policy gradient theorem. Specifically, we implement a variant of SPG called REINFORCE~\cite{williams1992, Huang2018LearningTP}. The agent parameters $\mathbf{\theta}$ are updated with gradient ascent so that actions with higher rewards are more probable to be sampled~\cite{williams1992}.
\begin{equation}
    \label{eq:Obective_grad}
        \nabla_{\mathbf{\theta}} J\left(\mathbf{\theta}\right) =  
        \nabla_{\mathbf{\theta}} \mathbb{E}\left[r_\text{prune}\right] +
        \nabla_{\mathbf{\theta}} \mathbb{E}\left[r_\text{retrain}\right]
\end{equation}

The first term in Eq.~\ref{eq:Obective_grad} has the Bernoulli policy 
$\pi_B\left(\mathbf{A}_\text{prune} | \mathbf{W}^\ell, \mathbf{P}, \mathbf{w}\right)$, while the second has the Normal policy $\pi_N\left(\mathbf{A}_\text{retrain} | \mathbf{W}^\ell, \mu, \mathbf{w}\right)$, where $\mathbf{W}^\ell$ are weights for layer to prune. 
Finding a closed-form solution for the expectation is not feasible, so it is 
approximated using $M$ samples of a Monte-Carlo gradient estimator with score function
~\cite{mohamed2019monte, sutton2018reinforcement}.
% \lukas{Might prune, since its stating a fact about MC which we already referred to: Monte-Carlo estimator is non-biased and sample-efficient that can be implemented even with one sample M = 1, however it has
% high variance Mohamed et al.~\cite{mohamed2019monte}.} 
% \ali{Done}
The gradient of our objective function is given by Eq.~\ref{eq:objective_final}.
%The whole process of pruning the full model is illustrated in Algorithm~\ref{alg:pruning_process}.
\begin{equation}
\small
    \label{eq:objective_final}
        \nabla_{\mathbf{\theta}} J\left(\mathbf{\theta}\right) \approx
        \sum_{j=1}^{M} \left[(R_\text{prune})_j \cdot \sum_{i=1}^{n} \frac{a_{ij}-p_{ij}}{p_{ij}(1-p_{ij})} \cdot
        \frac{\partial p_{ij}}{\partial \mathbf{w}} +
        ({R_\text{retrain}})_j \cdot \frac{a_{j}-\mu_{j}}{\sigma_j^2}  \cdot
        \frac{\partial \mu_{j}}{\partial \mathbf{w}}\right] 
\end{equation}

\section{Experimental Results}
\label{sec:experiments}
% \input{data/40_experiments/41_tabel_test.tex}
% \begin{figure}
%     \centering
%     \resizebox{1\linewidth}{!}{\input{data/40_experiments/45_Fig_5_2.tex}}
   
%     \label{fig:45_Fig_5_2}
% \end{figure}

% \begin{figure}
%     \centering
%     \scalebox{1.2}{\input{data/40_experiments/45_Fig_5_4.tex}}
    
%     \label{fig:45_Fig_5_4}
% \end{figure}
% \caption{Caption}
% \label{fig:my_label}
%% moved from method
Our experiments are conducted on the CIFAR-10~\cite{cifar} data set. One-time random splitting of the
50k images into 45k training and 5k evaluation is performed. Agent reward is evaluated on 5k
images. To ensure that our pruning method generalizes, the 10k images in the test data set are held separate and only used after the agent learns to prune a layer, to report actual
model accuracy. No training or reward evaluation is performed using the test data set.
As a baseline, ResNet-20 is trained from scratch as described in~\cite{he2015deep} until convergence with validation accuracy $92.0\%$, and test accuracy of $90.8\%$. After each pruning episode, the environment is retrained for a few epochs (8 w/o epochs learning) using mini-batch momentum SGD~\cite{qian1999momentum} with learning rate of 0.001, gamma 0.5, step size of 1900, batch size of 128, and $l2$ regularization. After learning to prune a layer, the model is fine-tuned for 150 epochs before moving to the next layer. The agent is also trained using mini-batch momentum SGD with fixed learning rate of 0.005 and batch size equal to the number of Monte-Carlo samples $M=5$. \\
% \manoj{fine tune epochs missing }

% \begin{table}[ht]
% \caption[Pruning configurations]{Pruning configurations, four configurations of two pruning strategies depending on pruning direction and number of layer to prune \manoj{Is this table important?}}
% \centering
% \resizebox{0.8\textwidth}{!}{
% \begin{tabular}{l|ccccc}
% \toprule
% \textbf{Configuration}
% & \textbf{Deep-to-Shallow} & \textbf{Shallow-to-Deep} & \textbf{Layer-wise} & \textbf{Residual-wise} \\
% \midrule
% \midrule
% D2S\_Rsdl & \cmark & - & - & \cmark\\ 
% D2S\_layer & \cmark & - & \cmark & -\\
% S2D\_Rsdl & - & \cmark & - & \cmark \\
% S2D\_layer & - & \cmark & \cmark& -\\

% \bottomrule
% \end{tabular}}
% %\vspace*{-2mm}
% \label{tab:related_work}
% \end{table}

    \subsection{Design Space Exploration}
    \subsubsection{Exploration of Efficient Pruning Order:}
\begin{figure}[t]
\centering
%\hspace{0.1\textwidth}
\captionsetup[subfigure]{justification=centering, font=scriptsize,labelfont=scriptsize}
\begin{subfigure}{\linewidth}    
\hspace{7ex} \begin{tikzpicture}	
	\begin{axis}[
		name=legend,
		width=6cm,
		height=2cm,
		hide axis,
        legend style={at={(0.05,0)},anchor=west, legend columns=-1, draw = none, nodes={scale=0.75, transform shape}, column sep=2pt},
		ymin=0,
        ymax=1,
        xmin=0,
        xmax=1,
		]
		\addlegendimage{thick, color= orange,solid,mark=star, mark size=2pt}
       	\addlegendentry{Accuracy Drop};
       	\addlegendimage{thick, color= red,solid,mark=square, mark size=1pt}
       	\addlegendentry{Sparsity Ratio};
       	\addlegendimage{line width=2pt, color=blue!60, solid, mark=square, only marks, mark size=1pt}
       	\addlegendentry{\# Remaining filters};
       	\addlegendimage{line width=2pt, color=green!80,mark=square,only marks, mark size=1pt}
       	\addlegendentry{\# Pruned filters};
	\end{axis}
\end{tikzpicture}
     \label{fig:my_label}
\end{subfigure}\\
  \begin{subfigure}[b]{0.5\linewidth}
    \pgfplotstableread{
        
    layer sparsity_ratio left_filters pruned_filters acc_drop
    1	0	16	0	0
    2	0.0080688542227	1	15	0.2
    3	0.01664201183432	1	15	0.2
    4	0.025215169445939	1	15	0.5
    5	0.033788327057558	1	15	0.4
    6	0.042361484669177	1	15	0.6
    7	0.050833781603012	4	12	0.9
    8	0.066837009144702	9	23	1.2
    9	0.097935718128026	11	21	0.899999999999992
    10	0.126075847229693	17	15	1.2
    11	0.149072081764389	20	12	1.4
    12	0.16601667563206	26	6	1.2
    13	0.172471759010221	32	0	1.3
    14	0.180002689618074	57	7	1.5
    15	0.20656266810113	58	6	1.1
    16	0.238972565895643	54	10	1.5
    17	0.278644432490586	54	10	1.6
    18	0.323762775685853	51	13	1.8
    19	0.442610274341044	11	53	1.8

    }\testdata
\resizebox{1\textwidth}{!}{
\begin{tikzpicture}

    \begin{axis}[
    %legend style={nodes={scale=0.6, transform shape}}, 
        %height = 0.5\textwidth,
        %width= 0.78\textwidth,
        height=2cm,
        width=6cm,
        ybar stacked,
        bar width = 0.2cm,
        scale only axis,
        %legend pos=north west,
        xmin=-1,xmax=19,
        ymin=0, ymax=80,
        axis y line*=left,
        grid = major,
        xlabel=Index of Layer,
        ylabel style = {align=center, font=\tiny, yshift=-4.5ex},
        xlabel style = {font=\tiny,yshift=1ex},
        ylabel={Number of Filters},
        xtick=data,
        yticklabel style = {font=\tiny},
        %legend style={
            %cells={anchor=west},
            %legend pos=north west,
        %},
        reverse legend=true,
        xticklabels from table={\testdata}{layer},
        xticklabel style={align=center, font=\tiny, rotate=45},
    ]

        \addplot [fill=blue!60]
            table [y=left_filters, meta=layer, x expr=\coordindex]
                {\testdata};
                    %\addlegendentry{Number of left filters}
        \addplot [fill=green!80]
            table [y=pruned_filters, meta=layer, x expr=\coordindex]
                {\testdata};
                    %\addlegendentry{Number of pruned filters}
	\node[align=center] (N3) at (axis cs: 3.5,65) {\tiny Shallow-to-Deep};
	\draw[->] (axis cs: 1,60) -- (axis cs: 6,60);
    \end{axis}
    
\begin{axis}[
%legend style={nodes={scale=0.6, transform shape}}, 
%legend pos = inner north east,
% height = 0.5\textwidth,
%         width= 0.78\textwidth,
height=2cm,
width=6cm,
scale only axis,
xmin=0,xmax=20,
ymin=0, ymax=2.5,
axis y line*=right,
ylabel near ticks, yticklabel pos=right,
yticklabel style = {font=\tiny},
axis x line=none,
xlabel style = {font=\tiny},
ylabel style = {align=center, font=\tiny},
%ylabel={Accuracy drop[\%]},
%legend style={
%            cells={anchor=west}
%            }
]
\addplot [ thick, color= orange,solid,mark=star, mark size=2pt]table [x=layer, y=acc_drop, col sep=comma] {Fig_5_1.csv};
\label{plot_acc_drop}

%\addlegendentry{Accuracy drop}

\addplot [ thick, color= red,solid,mark=square, mark size=1pt]table [x=layer, y=sparsity_ratio, col sep=comma] {Fig_5_1.csv};
%% \label{pgfplots:plot2}

%\addlegendentry{Model Sparsity ratio}

\end{axis}    

\end{tikzpicture}}
    \vspace{-3.5ex}
    \caption{ Forwards (Layer-wise)}\label{subfig:layerorder_in2out_layer}
\end{subfigure}%% 
    %\hspace{0.2\textwidth}
  \begin{subfigure}[b]{0.5\linewidth}
    \pgfplotstableread{
        
    layer left_filters pruned_filters acc_drop sparsity_ratio
    1	16	0	1.8	0
    2	11	5	1.8	0.002689618074233
    3	15	1	1.5	0.005749058633674
    4	13	3	1.59999999999999	0.007799892415277
    5	12	4	1.5	0.011161915008069
    6	10	6	1.59999999999999	0.015734265734266
    7	15	1	1.5	0.019298009682625
    8	21	11	1.61	0.025921194190425
    9	28	4	1.7	0.040579612694997
    10	21	11	1.7	0.05523803119957
    11	24	8	1.63	0.072720548682087
    12	24	8	1.55	0.087782409897795
    13	21	11	1.63	0.105264927380312
    14	40	24	1.59999999999999	0.145878160301237
    15	42	22	1.09999999999999	0.227104626143088
    16	20	44	0.799999999999997	0.336572081764389
    17	7	57	0.679999999999993	0.469573695535234
    18	1	63	0.629999999999995	0.607046799354492
    19	1	63	0.599999999999994	0.744721624529317

    }\testdata
\resizebox{1\textwidth}{!}{
\begin{tikzpicture}
    \begin{axis}[
    legend style={nodes={scale=0.6, transform shape}}, 
    % height = 0.5\textwidth,
    %     width= 0.78\textwidth,
    height=2cm,
    width=6cm,
    ybar stacked,
    bar width = 0.2cm,
scale only axis,
legend pos=north west,
xmin=-1,xmax=19,
ymin=0, ymax=80,
axis y line*=left,
grid = major,
xlabel=Index of Layer,
ylabel style = {align=center, font=\tiny, yshift=-4.5ex},
xlabel style = {font=\tiny,yshift=1ex},
yticklabel style = {font=\tiny},
        xtick=data,
        legend style={
            cells={anchor=west},
            legend pos=north west,
        },
        reverse legend=true,
        xticklabels from table={\testdata}{layer},
        xticklabel style={align=center, font=\tiny, rotate=45},
    ]
        \addplot [fill=blue!60]
            table [y=left_filters, meta=layer, x expr=\coordindex]
                {\testdata};
                    %\addlegendentry{Number of left filters}
        \addplot [fill=green!80]
            table [y=pruned_filters, meta=layer, x expr=\coordindex]
                {\testdata};
                    %\addlegendentry{Number of pruned filters}
	\node[align=center] (N3) at (axis cs: 3.5,65) {\tiny Deep-to-Shallow};
	\draw[<-] (axis cs: 1,60) -- (axis cs: 6,60);
    \end{axis}

\begin{axis}[
legend pos = north east,
legend style={nodes={scale=1.5, transform shape}}, 
% height = 0.5\textwidth,
%         width= 0.78\textwidth,
height=2cm,
width=6cm,
scale only axis,
xmin=0,xmax=20,
ymin=0, ymax=2.5,
axis y line*=right,
ylabel near ticks, yticklabel pos=right,
yticklabel style = {font=\tiny},
axis x line=none,
ylabel style = {align=center, font=\tiny},
xlabel style = {font=\tiny},
ylabel={Accuracy drop},
legend style={
            cells={anchor=west}
            }
]
\addplot [ thick, color= orange,solid,mark=star, mark size=2pt]table [x=layer, y=acc_drop, col sep=comma] {Fig_5_3.csv};
%% \label{pgfplots:plot2}

%\addlegendentry{Accuracy drop}

\addplot [ thick, color= red,solid,mark=square, mark size=1pt]table [x=layer, y=sparsity_ratio, col sep=comma] {Fig_5_3.csv};
%% \label{pgfplots:plot2}

%\addlegendentry{Model Sparsity ratio}

\end{axis}

\end{tikzpicture}}
    \vspace{-3.5ex}
    \caption{Backwards (Layer-wise)}
    \label{subfig:layerorder_out2in_layer}
     % \manoj{sudden drop in accuracy ? why?}
  \end{subfigure} \\
  \begin{subfigure}[b]{0.5\linewidth}
    \pgfplotstableread{
        
    layer sparsity_ratio left_filters pruned_filters acc_drop 
	1	0	16	0	0
    2	0.0080688542227	1	15	0.420000000000002
    3	0.016608391608392	2	14	0.420000000000002
    4	0.025013448090371	3	13	0.649999999999992
    5	0.033216783216783	4	12	0.649999999999992
    6	0.041420118343195	3	13	0.909999999999997
    7	0.04942173211404	6	10	0.909999999999997
    8	0.064416352877891	11	21	1
    9	0.093296126949973	15	17	1
    10	0.114611350188273	26	6	1.09999999999999
    11	0.127185314685315	25	7	1.09999999999999
    12	0.143121301775148	22	10	1.39999999999999
    13	0.157577998924153	27	5	1.39999999999999
    14	0.179229424421732	52	12	1.66
    15	0.222532275416891	54	10	1.66
    16	0.265835126412049	52	12	1.61999999999999
    17	0.317879236148467	49	15	1.61999999999999
    18	0.376512910166756	48	16	1.67999999999999
    19	0.456125605164067	36	28	1.67999999999999
    }\testdata
\resizebox{1\textwidth}{!}{
\begin{tikzpicture}
    \begin{axis}[
    legend style={nodes={scale=0.6, transform shape}}, 
    height = 2cm,
    width= 6cm,
    ybar stacked,
    bar width = 0.2cm,
scale only axis,
legend pos=north west,
xmin=-1,xmax=19,
grid = major,
ymin=0, ymax=80,
axis y line*=left,
xlabel=Index of Layer,
ylabel style = {align=center, font=\tiny, yshift=-4.5ex},
        xlabel style = {font=\tiny,yshift=1ex},
        ylabel={Number of Filters},
        xtick=data,
        yticklabel style = {font=\tiny},
        legend style={
            cells={anchor=west},
            legend pos=north west,
        },
        reverse legend=true,
        xticklabels from table={\testdata}{layer},
        xticklabel style={align=center, font=\tiny, rotate=45},
    ]
        \addplot [fill=blue!60]
            table [y=left_filters, meta=layer, x expr=\coordindex]
                {\testdata};
                    %\addlegendentry{Number of left filters}
        \addplot [fill=green!80]
            table [y=pruned_filters, meta=layer, x expr=\coordindex]
                {\testdata};
                    %\addlegendentry{Number of pruned filters}
	\node[align=center] (N3) at (axis cs: 3.5,65) {\tiny Shallow-to-Deep};
	\draw[->] (axis cs: 1,60) -- (axis cs: 6,60);
    \end{axis}

\begin{axis}[
legend pos =  north east,
legend style={nodes={scale=0.6, transform shape}}, 
height = 2cm,
width= 6cm,
scale only axis,
xmin=0,xmax=20,
ymin=0, ymax=2.5,
axis y line*=right,
ylabel near ticks, yticklabel pos=right,
axis x line=none,
ylabel style = {align=center, font=\tiny},
yticklabel style = {font=\tiny},
legend style={
            cells={anchor=west}
            }
]
\addplot [ thick, color= orange,solid,mark=star, mark size=2pt]table [x=layer, y=acc_drop, col sep=comma] {Fig_5_2.csv};
%% \label{pgfplots:plot2}

%\addlegendentry{Accuracy drop}

\addplot [ thick, color= red,solid,mark=square, mark size=1pt]table [x=layer, y=sparsity_ratio, col sep=comma] {Fig_5_2.csv};
%% \label{pgfplots:plot2}

%\addlegendentry{Model Sparsity ratio}

\end{axis}
   
\end{tikzpicture}}
    \vspace{-3.5ex}
    \caption{Forwards (Residual block-wise)}\label{subfig:layerorder_in2out_module}
  \end{subfigure}%%
  %\hspace{0.2\textwidth}
  \begin{subfigure}[b]{0.5\linewidth}
    \pgfplotstableread{
        
    layer sparsity_ratio left_filters pruned_filters acc_drop 
	1	0	16	0	1.5
    2	0.002689618074233	11	5	1.5
    3	0.007228348574502	11	5	1.5
    4	0.012506724045186	9	7	1.5
    5	0.016877353415815	14	2	1.5
    6	0.020306616460463	11	5	1.53999999999999
    7	0.023366057019903	15	1	1.53999999999999
    8	0.027467724583109	26	6	1.39999999999999
    9	0.040041689080151	25	7	1.39999999999999
    10	0.059339698762776	18	14	1.64
    11	0.078032544378698	26	6	1.64
    12	0.092354760623991	23	9	1.52
    13	0.107450242065627	25	7	1.52
    14	0.135960193652501	48	16	0.899999999999992
    15	0.204276492738031	43	21	0.899999999999992
    16	0.307288864981173	24	40	0.799999999999997
    17	0.433700914470145	14	50	0.799999999999997
    18	0.568114577729962	7	57	0.599999999999994
    19	0.704410973641743	6	58	0.599999999999994

    }\testdata
\resizebox{1\textwidth}{!}{
\begin{tikzpicture}
    \begin{axis}[
    legend style={nodes={scale=0.6, transform shape}}, 
    height = 2cm,
        width= 6cm,
        ybar stacked,
        bar width = 0.2cm,
scale only axis,
legend pos=north west,
grid = major,
xmin=-1,xmax=19,
ymin=0, ymax=80,
axis y line*=left,
xlabel=Index of Layer,
ylabel style = {align=center, font=\tiny, yshift=-4.5ex},
        xlabel style = {font=\tiny,yshift=1ex},
        xtick=data,
                yticklabel style = {font=\tiny},
        legend style={
            cells={anchor=west},
            legend pos=north west,
        },
        reverse legend=true,
        xticklabels from table={\testdata}{layer},
        xticklabel style={align=center, font=\tiny, rotate=45},
    ]
        \addplot [fill=blue!60]
            table [y=left_filters, meta=layer, x expr=\coordindex]
                {\testdata};
                    %\addlegendentry{Number of left filters}
        \addplot [fill=green!80]
            table [y=pruned_filters, meta=layer, x expr=\coordindex]
                {\testdata};
                    %\addlegendentry{Number of pruned filters}
	\node[align=center] (N3) at (axis cs: 3.5,65) {\tiny Deep-to-Shallow};
	\draw[<-] (axis cs: 1,60) -- (axis cs: 6,60);
    \end{axis}

\begin{axis}[
legend pos = north east,
legend style={nodes={scale=0.6, transform shape}}, 
height = 2cm,
width= 6cm,
scale only axis,
xmin=0,xmax=20,
ymin=0, ymax=2.5,
axis y line*=right,
ylabel near ticks, yticklabel pos=right,
axis x line=none,
ylabel style = {align=center, font=\tiny},
ylabel={Accuracy drop},
yticklabel style = {font=\tiny},
legend style={
            cells={anchor=west}
            }
]
\addplot [ thick, color= orange,solid,mark=star, mark size=2pt]table [x=layer, y=acc_drop, col sep=comma] {Fig_5_2.csv};
%% \label{pgfplots:plot2}

%\addlegendentry{Accuracy drop}

\addplot [ thick, color= red,solid,mark=square, mark size=1pt]table [x=layer, y=sparsity_ratio, col sep=comma] {Fig_5_2.csv};
%% \label{pgfplots:plot2}

%\addlegendentry{Model Sparsity ratio}

\end{axis}
   
\end{tikzpicture}}
    \vspace{-3.5ex}
    \caption{Backwards (Residual block-wise)}\label{subfig:layerorder_out2in_module}
  \end{subfigure}\\
  \begin{subfigure}[b]{\linewidth}
    \pgfplotstableread{
        
    layer epochs_with_learning pruned_filters_with_epochs_learning pruned_filters_without_epochs_learning epochs_without_learning 
    1     0     0   0   0
    2	 2.517	0	5	3.200
    3	 2.243	0	1	2.934
    4	 2.981	0	3	3.105
    5	 2.700	0	4	2.874
    6	 1.666	0	6	3.072
    7	 2.420	0	1	4.680
    8	 1.509	3	11	3.200
    9	 1.212	5	4	1.920
    10 1.001	6	11	2.520
    11 	1.100	7	8	2.800
    12 	1.485	7	8	4.600
    13 	0.942 	16	11	2.600
    14 	2.176	17	24	8.000
    15 	4.074	42	22	2.760
    16 	1.945	49	44	5.920
    17 	2.144	55	57	2.000
    18 	1.560	47	63	2.040
    19	1.575	57	63	2.080
    }\testdata

\begin{tikzpicture}

\begin{axis}[
ybar,
legend pos=north east,
legend style={nodes={scale=0.6, transform shape}, at={(-1.6,.3)},anchor=north},
legend image code/.code={
        \draw [#1] (0cm,-0.1cm) rectangle (0.2cm,0.1cm); },
legend entries = {Filters pruned without epoch learning, Filters pruned with epoch learning},
bar width = 0.15cm,
scale only axis,
grid = major,
height = 2.0cm,
width= 10.7cm,
xmin=-1,xmax=19,
ymin=0, ymax=70,
axis y line*=left,
yticklabel style = {font=\tiny},
xticklabel style = {font=\tiny},
xlabel=Index of Layer,
xlabel style = {align=center, font=\tiny, yshift=1ex},
ylabel style = {align=center, font=\tiny, yshift=-5ex},
ylabel={Number of Filters},
        xtick=data,
        legend style={
            cells={anchor=west},
            legend pos=north west,
        },
        reverse legend=true,
        xticklabels from table={\testdata}{layer},
        xticklabel style={text width=2cm,align=center},
    ]
        \addplot [fill=Blue3, bar shift=.08cm]
            table [y=pruned_filters_without_epochs_learning, meta=layer, x expr=\coordindex]
                {\testdata};

        \addplot [fill=Green3, bar shift=-.08cm]
            table [y=pruned_filters_with_epochs_learning, meta=layer, x expr=\coordindex, bar shift=-2pt]
                {\testdata};
                    %% \label{plot2}
\end{axis}
    
\begin{axis}[
legend pos = north east,
legend style={nodes={scale=0.6, transform shape}},
height = 2.0cm,
width= 10.7cm,
scale only axis,
xmin=0,xmax=20,
ymin=0, ymax=9.5,
axis y line*=right,
ylabel near ticks, yticklabel pos=right,
yticklabel style = {font=\tiny},
axis x line=none,
ylabel style = {align=center, font=\tiny},
ylabel={Finetune Epochs $\cdot 10^3$},
legend style={at={(0.39,0.97)},anchor=north west}
]

\addplot [thick, color= Blue1,solid,mark=triangle, mark size=2pt]table [x=layer, y=epochs_with_learning, col sep=comma] {Fig_5_7_b.csv};

\addlegendentry{Epochs with learning}

\addplot [thick, color= Blue2,solid,mark=diamond, mark size=2pt]table [x=layer, y=epochs_without_learning, col sep=comma] {Fig_5_7_b.csv};

\addlegendentry{Epochs without learning}

\end{axis}
   
\end{tikzpicture}
    \vspace{-3.5ex}
    \caption{Layer-wise reduction in fine-tuning epochs with and without learning.} 
    % \manoj{Bound 2 or Bound 1?}
    \label{fig:retrain_epochs}
\end{subfigure}
  \caption{Exploration of filter pruning order and epoch-learning effect.}
  \label{fig:Fig3_layer_wise_pruning} 
\end{figure}
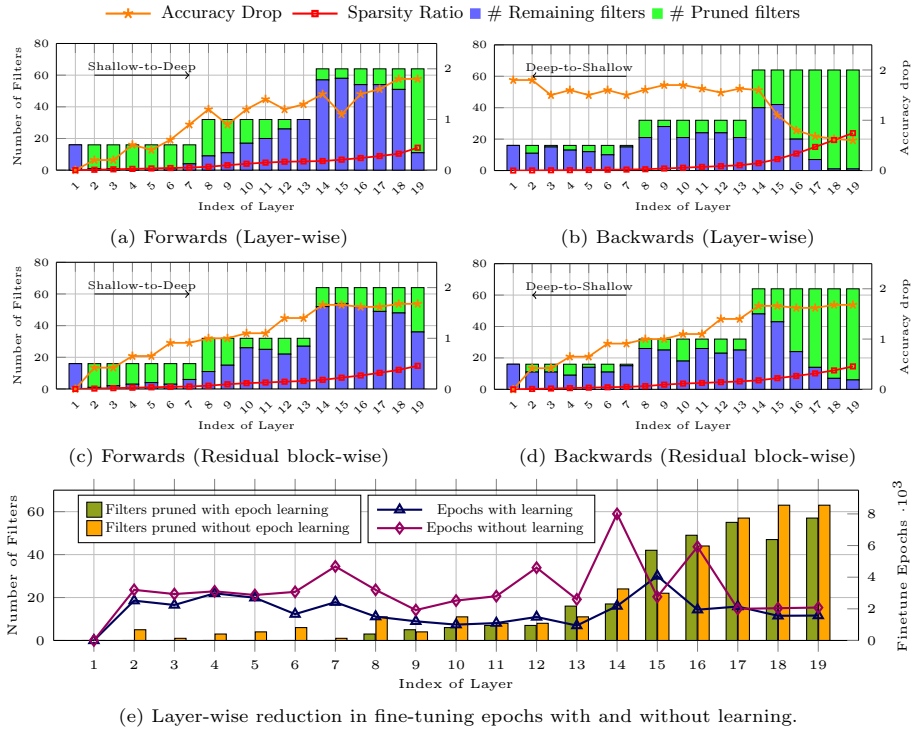

% \ali{check consistency of using block-wise\/residual-wise}
We investigate four different strategies based on the pruning order and the agent's capability to prune layers simultaneously (layerwise or blockwise). We exclude the first convolutional layer since pruning it offers insigniﬁcant compression benefits, while damaging the learning ability of the model. When pruning a full residual block, we preserve the element-wise summation by zero-padding the output channels of the second layer in a residual block to restore the original number of output channels, such that the order of pruned channels is preserved. Fig.~\ref{fig:Fig3_layer_wise_pruning} shows results of pruning ResNet-20 with loss bound $b$ of 2\%.% Differently speaking $2\%$ accuracy degradation caused by the agent is accepted.%The stacked bars show the number of pruned and remaining filters per layer, orange line with markers show accuracy drop after pruning and fine-tune per layer, red line shows total model sparsity level on the right axis.
% Configuration 1:
In Fig.~\ref{subfig:layerorder_in2out_layer}, we perform layer-wise pruning following the forward pruning order ($\mathtt{conv2\_1\_1} \rightarrow \mathtt{conv4\_3\_2}$). The agent starts pruning initial layers aggressively and struggles to find redundant filters in the deep layers (indicated by large blue bars). 
In Fig.~\ref{subfig:layerorder_out2in_layer}, we perform a similar layer wise pruning analysis for the backward pruning order ($\mathtt{conv2\_1\_1} \leftarrow \mathtt{conv4\_3\_2}$). From Tab.~\ref{tab:my_label}, we observe that the backward pruning order results in higher CR with lower accuracy degradation (0.9\%). 
In Fig.~\ref{subfig:layerorder_in2out_module} and~\ref{subfig:layerorder_out2in_module}, we perform block wise pruning allowing the agent to prune the entire residual block simultaneously. Similar to layer-wise pruning, we prune the residual blocks both in forward and backward orders. Lower compression ratio is observed when compared to layer-wise pruning, see also Tab.~\ref{tab:my_label}. Thus, we prune layer-wise in backward order as it results in lower accuracy degradation and high CR for the subsequent experiments. 

\begin{table}[t]
    \centering
    \caption{Evaluating various configurations for L2PF to analyze the influence of exploration granularity, pruning order, accuracy bound w.r.t. prediction accuracy and compression ratio.}
    \resizebox{0.65\textwidth}{!}{
\begin{tabular}{l|ccccc}
    \toprule
    \multirow{2}{*}{Configuration} & \multirow{1}{*}{Pruning} & \multirow{1}{*}{Bound} & \multirow{1}{*}{Learnable}&\multirow{1}{*}{Acc} & \multirow{1}{*}{CR}\\
    &\multirow{1}{*}{Order}&\multirow{1}{*}{[$\%$]}&\multirow{1}{*}{Epochs}&\multirow{1}{*}{[$\%$]}&\multirow{1}{*}{[$\times$]}\\
    \midrule
    \midrule
    ResNet-20~\cite{he2015deep}      & - & - & - & 90.8 & 1.00 \\
    \midrule
    L2PF (Block-wise) & Forwards & 2.0 &\xmark& 89.9 (-0.9) & 1.84 \\
    L2PF (Layer-wise) & Forwards & 2.0 &\xmark& 89.6 (-1.2) & 1.79 \\
    \midrule
    L2PF (Block-wise) & Backwards & 2.0 &\xmark& 89.5 (-1.3) & 3.38 \\
    L2PF (\textbf{Layer-wise}) & \textbf{Backwards} & 2.0 &\xmark& \textbf{89.9} (-0.9) & \textbf{3.90} \\
    \midrule
    L2PF (Layer-wise) & Backwards & 1.0 &\xmark& 90.2 (-0.6) & 2.52 \\
    L2PF (\textbf{Layer-wise}) & \textbf{Backwards} & \textbf{2.0} &\xmark& \textbf{89.9} (-0.9) & \textbf{3.90} \\
    L2PF (Layer-wise) & Backwards & 3.0 &\xmark& 89.2 (-1.0) & 4.53 \\
    L2PF (Layer-wise) & Backwards & 4.0 &\xmark& 88.5 (-2.3) & 7.23 \\
    \midrule
    L2PF (Layer-wise) & Backwards & 2.0 &\xmark& 89.9 (-0.9) & 3.90 \\
    \textbf{L2PF (Layer-wise)} & \textbf{Backwards} & \textbf{2.0} &\cmark& \textbf{89.9} (-0.9) & \textbf{3.84} \\    
    \bottomrule
\end{tabular}}
    \label{tab:my_label}
\end{table}

\begin{table}[b!]
    \centering
    \begin{tabular}{c|c|cccc}
    \toprule
        \textbf{Input} & \textbf{ResNet-20} & \multicolumn{4}{c}{\textbf{Learning to Prune 
        Faster (Backwards)}}\\
        \textbf{image} & \textbf{unpruned} & $\mathtt{conv4\_3\_2} \rightarrow$ & $\mathtt{conv3\_3\_2} \rightarrow$ & $\mathtt{conv2\_3\_2} \rightarrow$ & $\mathtt{conv2\_1\_2}$ \\ 
        \midrule
        \midrule
        \includegraphics[width=0.11\linewidth]{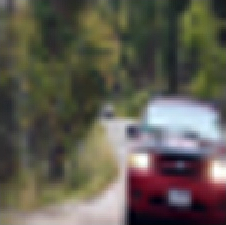} & \includegraphics[width=0.11\linewidth]{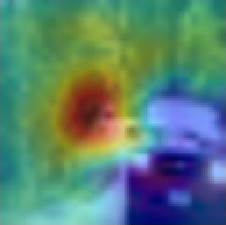} & \includegraphics[width=0.11\linewidth]{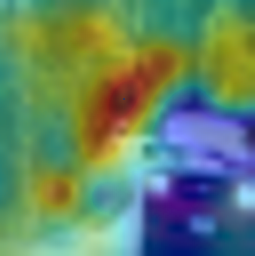} & \includegraphics[width=0.11\linewidth]{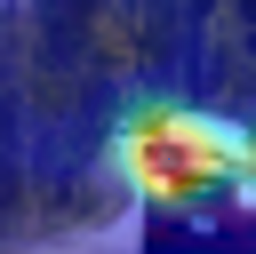} & \includegraphics[width=0.11\linewidth]{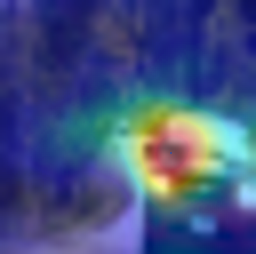} & \includegraphics[width=0.11\linewidth]{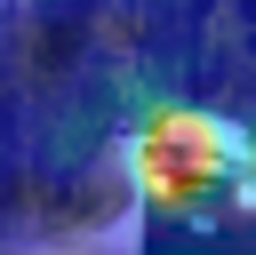} \\
        raw & \emph{deer}(0.53) & \emph{car}(0.99) $\rightarrow$ & \emph{car}(0.99) $\rightarrow$ & \emph{car}(0.99) $\rightarrow$ & \emph{car}(0.88) \\ 
        \midrule
        \includegraphics[width=0.11\linewidth]{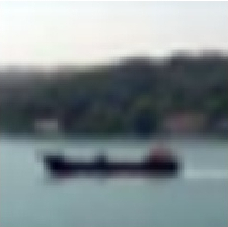} & \includegraphics[width=0.11\linewidth]{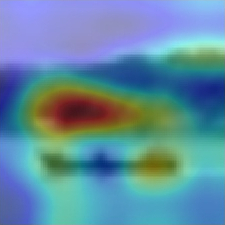} & \includegraphics[width=0.11\linewidth]{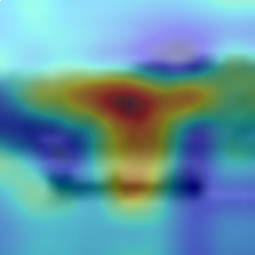} & \includegraphics[width=0.11\linewidth]{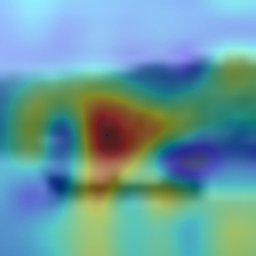} & \includegraphics[width=0.11\linewidth]{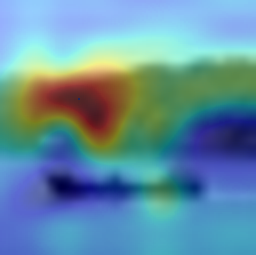} & \includegraphics[width=0.11\linewidth]{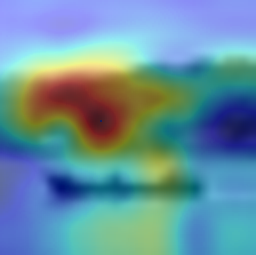} \\
        raw & \emph{ship}(0.99) & \emph{ship}(0.51) $\rightarrow$ & \emph{ship}(0.98) $\rightarrow$ & \emph{ship}(0.81) $\rightarrow$ & \emph{ship}(0.99) \\ 
        \midrule
        \includegraphics[width=0.11\linewidth]{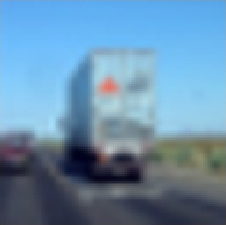} & \includegraphics[width=0.11\linewidth]{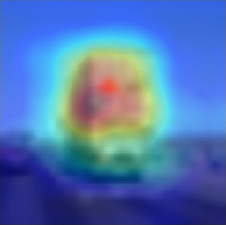} & \includegraphics[width=0.11\linewidth]{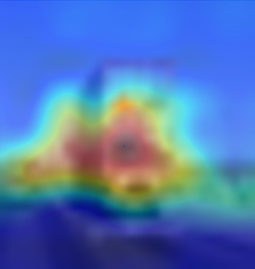} & \includegraphics[width=0.11\linewidth]{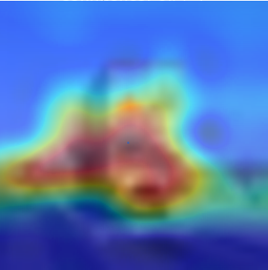} & \includegraphics[width=0.11\linewidth]{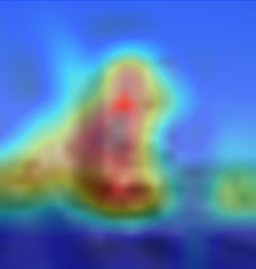} & \includegraphics[width=0.11\linewidth]{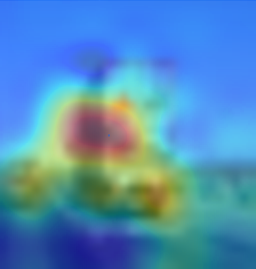} \\
        raw & \emph{truck}(0.99) & \emph{truck}(0.98) $\rightarrow$ & \emph{truck}(0.77) $\rightarrow$ & \emph{truck}(0.62) $\rightarrow$ & \emph{truck}(0.67) \\
        \bottomrule
    \end{tabular}
    \caption{CAM visualization for three examples images from the validation dataset. Each column shows the CAM output after pruning, using backwards pruning order before model fine-tuning.}
    \label{fig:CAM_visualization}
\end{table}

\subsubsection{Effect of Accuracy Bound on Compression Ratio:}
In Tab.~\ref{tab:my_label}, we also evaluate the impact of the prediction accuracy and compression ratio by varying the agent's loss bound $b$. As we increase $b$, we obtain higher CR with lower prediction accuracy after fine-tuning. We choose $b$ as 2\% to maintain a trade-off between accuracy degradation and CR. 

\subsubsection{Effect of Accuracy and Pruning Rate on Exploration Time:}
Previous experiments were conducted with fine-tuning epochs set manually to 8 at each exploration step. We allow the agent to decide the amount of fine-tuning time required to evaluate the pruning strategy based on the retrain epoch reward presented in Eq.~\ref{eq:retrain_reward}. Fig.~\ref{fig:retrain_epochs} shows a comparison of number of fine-tuning epochs required to decide the pruning strategy for each layer. Pruning with epochs learning achieves  $1.71 \times$ speedup in search time with a slight reduction in compression ratio, see Tab.~\ref{tab:my_label}. 

    \subsection{Class Activation Maps}
    %CAM introduction
The discrete action space proposed by Huang \etal~\cite{Huang2018LearningTP} and applied in L2PF (described in Sec.\ref{sec:action_space}) allows the integration of class activation mapping (CAM)~\cite{zhou2015learning} into the design process. CAM allows the visualization of regions of interest (RoI) in an input image to identify the corresponding prediction label. Regions with red color denote the part with higher interest for CNN model and blue denotes regions with less importance w.r.t. the target label. Tab.~\ref{fig:CAM_visualization} shows three exemplary CAMs for the learned features of vanilla ResNet-20 and the influence of L2PF pruning (backwards) on the learned features and thus the RoIs. The progression of discriminative regions of classes can be compared across pruning steps.

%We apply CAM (Class Activation Map)~\cite{zhou2015learning} to visualize the Regions of Interest (RoI) in the input image for the pruned model to identify the corresponding prediction label. In this way, progression of discriminative regions of class can be compared across pruning steps. Regions with red color denote the part with higher interest for CNN model and blue denotes regions with less importance w.r.t. the target label. We demonstrate the CAM visualization by randomly picking three example images from the validation set and observing its effect at different stages of pruning in Fig~\ref{fig:CAM_visualization}. 

%CAM interpretation (b)
In the first row, the vanilla ResNet-20 predicts the wrong class, \ie \emph{deer}. After pruning layer $\mathtt{conv3\_3\_2}$, the RoI shifts towards the \emph{trunk} of the \emph{car} indicating the correct class.
%CAM interpretation (c)
In the second row, the vanilla ResNet-20 predicts the \emph{ship} class. The agent tries to retain the prediction across different stages of pruning with high confidence.
%CAM interpretation (a)
In the third row, the vanilla ResNet-20 predicts the \emph{truck} class. Accordingly, the pruned model at different stages also predict a \emph{truck}. However, we can observe that the RoI becomes narrower indicating that the pruned model requires only few concentrated regions due to lower model capacity. 
% The confidence of the prediction is reduced from 99.9\% to 66.6\%.
We consider potential directions of our future work as follows: (1) Considering the CAM output as a state embedding instead of a weight matrix, making the pruning more feature-aware and interpretable. This would not possible with threshold based approaches like AMC~\cite{He2018AMCAF}, (2) Understanding, the impact of feature-aware pruning on model robustness~\cite{Ye_2019_ICCV}.
    \subsection{Comparison with the State-of-the-Art}
    % \begin{table*}[!ht]
% \centering
% \caption[Comparison with State of the Art]{Comparison of Accuracy and compression ratio of our method with manual pruning methods (uniform, shallow,
% deep) and~\cite{He2018AMCAF} on Plain-20}
%   \label{tab:sota}
%  \resizebox{0.8\textwidth}{!} {\begin{tabular}{llllll} 
%   \toprule
%     \textbf{Parameter} & \textbf{D2S\_layer\_b1} & \textbf{Uniform} &\textbf{Shallow} &\textbf{Deep} &\textbf{AutoML} \\
%     \midrule
%     {Accuracy[\%]}  &  90.22   &  89.7    &  89.2 & 88.3 & 90.2 \\
% {Compression\_Ratio} &  2.52\times & 1.35\times & 1.56\times & 1.27\times & 1.37\times \\
%     \bottomrule
%     %%$\multicolumn{3}{l}{$^{1}$} \\
%     \end{tabular}}
% \end{table*}

\begin{figure}[h!]
%\begin{floatrow}
%\ffigbox{%
\begin{minipage}[b]{0.5\textwidth}
\centering
  %\resizebox{0.5\textwidth}{!}{
\begin{tikzpicture}

\begin{axis}[
legend style={nodes={scale=0.7, transform shape}}, legend columns= 2
        % legend image post style={mark=*},
legend pos = outer north east,
grid =major,
height=3.5cm,
width=7cm,
xmin=1,xmax=19,
ymin=0, ymax=1.1,
xlabel=Layer Index, 
xlabel style ={align=center, font=\tiny, yshift=1.5ex
},
ylabel style={align=center, font=\tiny
,yshift=-5.5ex},
ylabel={Compression Ratio},
yticklabel style = {font=\tiny, rotate=90},
xticklabel style = {font=\tiny, rotate=45},
xtick={1,2,...,19},
legend style={at={(0.5,1)},anchor=south, draw=none}
]
% \addplot [thick, color= purple,solid,mark=o*, mark size=2pt]table [x=layer, y=AutoML, col sep=comma] {Fig_5_8_a.csv};

% \addplot [very thick, color= yellow,solid,mark=star, mark size=2pt]table [x=layer, y=uniform, col sep=comma] {Fig_5_8_a.csv};

% \addplot [thick, color= green,solid, mark=diamond*, mark size=2pt]table [x=layer, y=shallow, col sep=comma] {Fig_5_8_a.csv};

% \addplot [ thick, color= orange,solid,mark=square*, mark size=2pt]table [x=layer, y=deep, col sep=comma] {Fig_5_8_a.csv};

% \addplot [thick, color= red,solid, mark=+, mark size=2pt]table [x=layer, y=Layerwise, col sep=comma] {Fig_5_8_a.csv};

% \addplot [thick, color= blue,solid, mark=*, mark size=2pt]table [x=layer, y=Rsdlwise, col sep=comma] {Fig_5_8_a.csv};

% \legend{AutoML (90.2\%), uniform (89.7\%), shallow (89.2\%), deep (88.3\%), Layerwise (b=2\%-89.6\%),Rsdlwise(b=2\%-89.1\%)}

\addplot [thick, color= purple,dashed]table [x=layer, y=AMC-Plain20, col sep=comma] {Fig_5_8_a.csv};

%\addplot [thick, color= blue,solid]table [x=layer, y=AMC-ResNet20, col sep=comma] {Fig_5_8_a.csv};

\addplot [thick, color= gray,solid]table [x=layer, y=ForwardLayerwise, col sep=comma] {Fig_5_8_a.csv};

%\addplot [thick, color= black,dashdotted]table [x=layer, y=ForwardResidualwise, col sep=comma] {Fig_5_8_a.csv};

%\addplot [thick, color= gray,dashdotted]table [x=layer, y=BackwardResidualwise, col sep=comma] {Fig_5_8_a.csv};

% \addplot [thick, color= green, densely dotted]table [x=layer, y=BackwardLayerwiseEpochB1, col sep=comma] {Fig_5_8_a.csv};

\addplot [thick, color= teal, densely dotted]table [x=layer, y=ALFResNet20, col sep=comma] {Fig_5_8_a.csv};

\addplot [thick, color= blue]table [x=layer, y=BackwardLayerwiseEpochB2, col sep=comma] {Fig_5_8_a.csv};

\legend{AMC-Plain20 \cite{He2018AMCAF},%, AMC-ResNet20 (91.7\%) \cite{He2018AMCAF}, 
 L2P\cite{Huang2018LearningTP}, ALF-ResNet20\cite{Frickenstein2020ALFAL}, L2PF (Ours)}
%Backward Layer-wise (Ours)(b=2\%-89.9\%)
% ,Forward Residual-wise(b=2\%-89.1\%), Backward Residual-wise(b=2\%-89.5\%)

% Backward Layer-wise w/ epoch learning(b=1,\%-\%)

\end{axis}    
\end{tikzpicture}%}
  \vspace{-5ex}
      %\label{fig:fig_5_5_and_5_6}
%}%{%
  %\caption{Compression ratio per layer of our approach on ResNet-20 (deep to shallow layer-wise/residual block-wise and shallow to deep residual block-wise) compared to ~\cite{Huang2018LearningTP} on ResNet-20(shallow to deep layer-wise), and compared to~\cite{He2018AMCAF} on Plain-20 and ResNet-20.}%
%}
\par\vspace{0pt}
\end{minipage}%
%\vspace{-5ex}
\begin{minipage}[b]{0.5\textwidth}
%\capbtabbox{%
\centering
\resizebox{0.9\textwidth}{!}{
\begin{tabular}[b]{l|cccc}
    \toprule
    \multirow{2}{*}{\textbf{Configuration}} & \multirow{1}{*}{\textbf{Pruning}} &\multirow{1}{*}{\textbf{Acc}} & \multirow{1}{*}{\textbf{CR}}& \multirow{1}{*}{\textbf{Fine-tune}}\\
    & \multirow{1}{*}{\textbf{Type}}&\multirow{1}{*}{[$\%$]}&\multirow{1}{*}{[$\times$]}& \multirow{1}{*}{\textbf{[epochs]}}\\
    \midrule
    \midrule
    ResNet-20~\cite{he2015deep} &-& 90.8 & 1.00& -\\
    \midrule
    AMC-Plain20~\cite{He2018AMCAF} &RL-agent&  90.2 & 1.84&- \\
    ALF~\cite{he2015deep} &In-train& 89.4 & 3.99& -\\
    % AMC-ResNet20 & 91.7 & 1.53 \\
    L2P~\cite{Huang2018LearningTP} &RL-agent& 89.6  & 1.79& 60.3K\\
    \textbf{L2PF (Ours) }&\textbf{RL-agent}& \textbf{89.9}  & \textbf{3.84}&\textbf{35.2K} \\
    \bottomrule
\end{tabular}}
\par\vspace{0pt}
%}%{%
 % \caption{tab.res}%
%}
%\end{floatrow}
\end{minipage}
\caption{Comparing L2PF pruning statistics on ResNet-20 with State-of-the-Art.}
\label{fig:SoTA}
\end{figure}
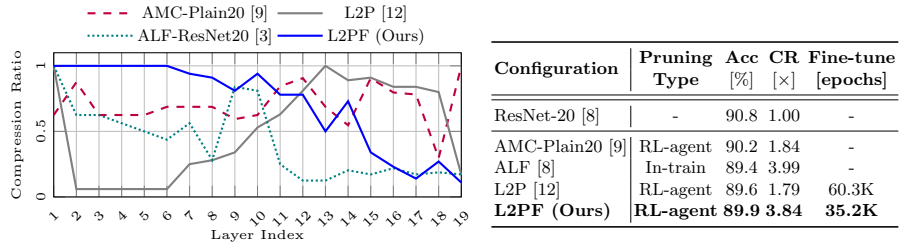

In this section, we compare the proposed L2PF with other RL-based state-of-the-art filter pruning works proposed in literature. 
In Fig.~\ref{fig:SoTA}, we compare our pruning configuration using layer-wise CR and final prediction accuracy with AMC~\cite{He2018AMCAF}, L2P~\cite{Huang2018LearningTP}, ALF~\cite{Frickenstein2020ALFAL}. We reimplemented L2P using forward pruning order with an accuracy bound $b$=2\% to obtain pruning results for ResNet-20. Compared to L2P, we obtain 0.3\% better prediction accuracy, 2.11$\times$ higher CR and 1.71$\times$ less fine-tune epochs. ALF and AMC do not require fine-tuning during the pruning process. Compared to AMC's pruning implementation for Plain-20, we obtain 2.08$\times$ higher CR with 0.3\% lower prediction accuracy. Compared to ALF, we achieve 0.5\% better accuracy with comparable CR. 
%%\manoj{Ali: May you read the section and fill the final reduction value}
\vspace{-3ex}
\vspace{-6.5ex}
\section{Conclusion}
\label{sec:conclusion}
In this work, we demonstrated an RL-based filter-wise pruning method which is both feature and time-aware. Our multi-task approach achieved high compression ratios, while minimizing the required GPU-hours and the accuracy degradation. The analysis on the sequence of layer-wise pruning led to the conclusion that backward (deep-to-shallow) pruning can surpass the existing state-of-the-art compression ratios, with minimal degradation in task accuracy.
%\alex{continuous and discrete action space, by Bernoulli and normal policies }\nael{too detailed for conclusion?}
Finally, we visually analyzed the effect of our pruning technique with the help of class activation maps to build a better understanding of our agent's pruning decisions. GPU-hours for CNN compression can have many negative consequences on development cycles, profitability and fast exploration. The GPU-hour-aware approach presented can help mitigate this impediment and achieve a competitive advantage in active research fields such as autonomous driving.

\iffalse

\begin{itemize}
    \item Location
    \item Pruning order
    \item exploration time
    \item good compression ratio, minimal degredation (backward) layerwise.
    \item visual undestanding
\end{itemize}

In this work, we build upon a learning-based pruning approach [9] by tackling
the challenge of choosing the optimal number of retraining epochs
3
for a po-
tential pruning solution. The decisions of our RL-based pruning agent are based
not  only  on  the  features  embedded  in  the  CNN’s  kernels,  but  also  on  the  re-
training potential of the layer. This results in feature-conscious decisions which
can  potentially  improve  the  robustness  of  the  resulting  network  against  mali-
cious  adversarial  attacks  and  reduce  the  overall  time  required  by  the  pruning
technique. The contributions of this paper can be summarized as follows:
–
A  multi-task  learning  approach  involving  a  reinforcement  learning  agent,
which learns a layer’s features and adequate retraining time concurrently.
–
Formalizing the design space exploration problem w.r.t. pruning effectiveness
and time-effort.
–
A  study  on  the  sequence  of  layer-wise  pruning  of  a  convolutional  neural
network.
\fi

%
\bibliographystyle{splncs04}
\bibliography{references}
\end{document}